\documentclass[letterpaper, 10 pt, conference]{ieeeconf}  

\IEEEoverridecommandlockouts                              
\usepackage[utf8]{inputenc}
\usepackage{graphicx}
\usepackage{amsmath, amssymb}
\DeclareMathOperator{\E}{\mathbb{E}}
\usepackage{bm}
\usepackage[keeplastbox]{flushend}
\usepackage{multirow}
\usepackage{hyperref}
\hypersetup{
    colorlinks,
    linkcolor={blue!30!black},
    citecolor={blue!30!black},
    urlcolor={white!10!black}
}
\usepackage{tabularx, booktabs, threeparttable, caption}
\newcolumntype{P}[1]{>{\centering\arraybackslash}p{#1}}
\newcolumntype{M}[1]{>{\centering\arraybackslash}m{#1}}

\usepackage{caption}
\usepackage{subcaption}

\usepackage[style=ieee, backend=biber, hyperref=true, url=false, isbn=false, doi=false, backref=false, minnames=1,maxnames=2]{biblatex}
\newcommand{\citet}[1]{\citeauthor*{#1} \cite{#1}}
\bibliography{other,references_Phuong,pps_deep,visual-motor-learning}

\newcommand{\norm}[1]{\left\lVert#1\right\rVert}
\graphicspath{{figures/}}
\usepackage{blindtext}
\usepackage{xcolor}

\usepackage{fancyhdr}
\fancypagestyle{firstpage}{
  \lhead{Preprint, submitted to \textit{IEEE Robotics and Automation Letters (RA-L) 2021} with \textit{International Conference on Robotics and Automation Conference Option (ICRA)} 2021}
}

\title{\LARGE \bf
Robotic self-representation improves manipulation skills and transfer learning 
}

\author{Phuong D.H. Nguyen$^{1}$, Manfred Eppe$^{1}$, Stefan Wermter$^{1}$
\thanks{$^{1}$Phuong D.H. Nguyen, Manfred Eppe and Stefan Wermter are with Department  of  Informatics, University of Hamburg, Hamburg, Germany
        {\tt\small \{pnguyen, eppe, wermter\}@informatik.uni-hamburg.de}}%
}

\begin{document}

\maketitle
\thispagestyle{firstpage}
\pagestyle{empty}

\begin{abstract}
Cognitive science suggests that the self-representation is critical for learning and problem-solving. However, there is a lack of computational methods that relate this claim to cognitively plausible robots and reinforcement learning. 
In this paper, we bridge this gap by developing a model that learns bidirectional action-effect associations to encode the representations of body schema and the peripersonal space from multisensory information, which is named multimodal BidAL.
Through three different robotic experiments, we demonstrate that this approach significantly stabilizes the learning-based problem-solving under noisy conditions and that it improves transfer learning of robotic manipulation skills.
\end{abstract}

\section{INTRODUCTION}

\begin{figure}[t]
   \centering
   \includegraphics[width=0.75\linewidth]{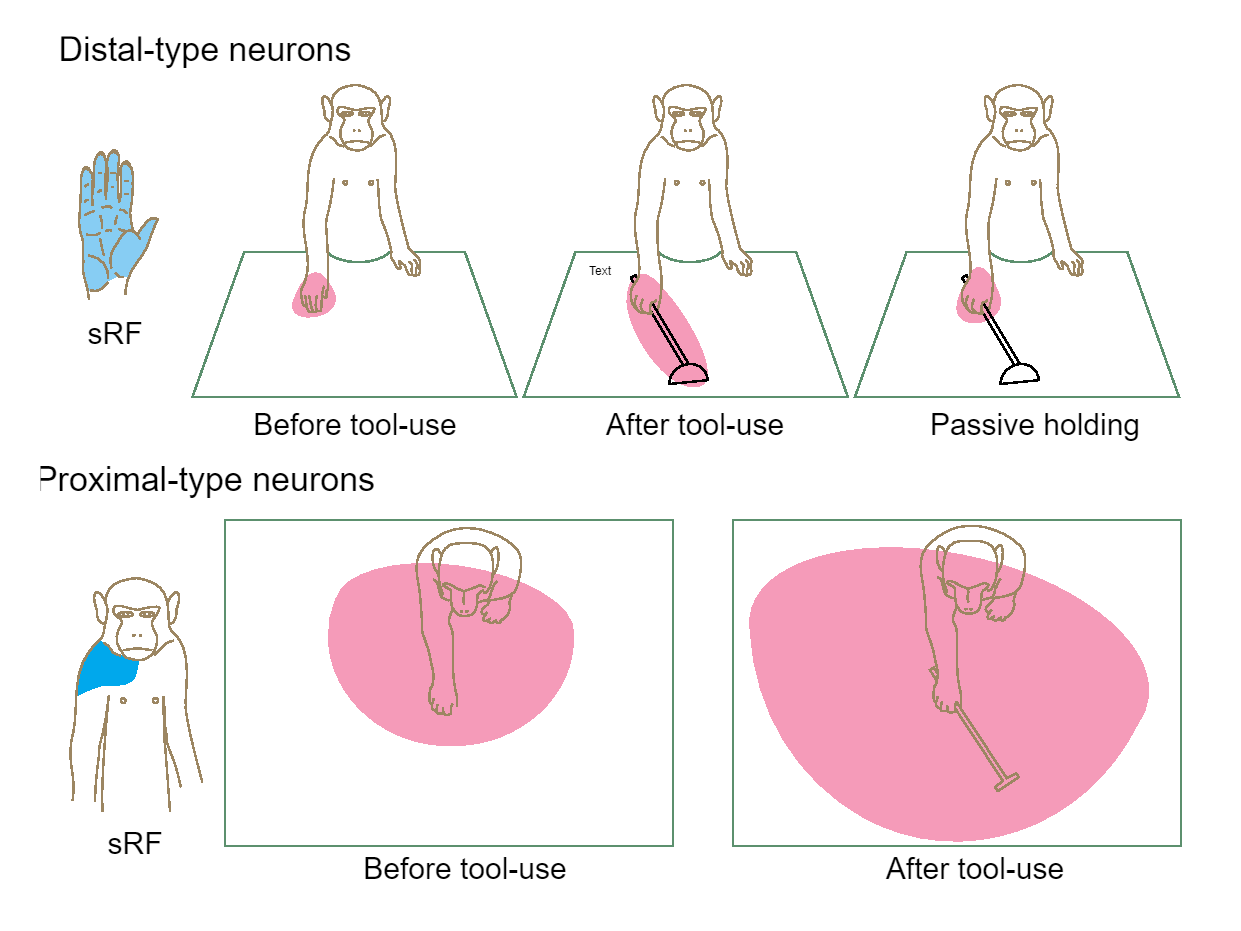}
   \caption{The body schema and peripersonal space representations during tool-use (from~\cite{Maravita2004Toolsschema}
   ).}
  \vspace{-0.75cm}
   \label{fig:bs-pps}
\end{figure}

Humans and other biological agents depend on appropriate representations of their body and their surroundings to facilitate their activities in safety and comfort. These representations, known as body schema and peripersonal space (PPS) (see 
Fig.~\ref{fig:bs-pps}), result from the integration of different sensorimotor modalities that are involved when physically interacting with the environment~\cite{head_sensory_1911,deVignemont2010BodyCons,clery_neuronal_2015,Serino2019PeripersonalSelf}.

Consequently, the body schema and peripersonal space are not innate but develop incrementally. 
This developmental process starts in infants through self-exploration and motor babbling, and continues later through goal-directed physical and social interactions~\cite{Rochat1995SpatialInfants,Bremner2012MultisensoryDevelopment,Corbetta2018TheCoordination}. 
Acquiring a body schema and a peripersonal space representation involves 
associative learning, a mechanism enabling infants to detect the sensorimotor contingencies in their environment. However, 
associative learning alone is not sufficient to explain the ability of humans to generate goal-directed actions. 
The conversion of the learned contingencies into goal-directed actions is related to another central ability that is only little investigated in computational methods: 
The ability to distinguish between self-caused body-schematic sensory effects and externally caused sensory effects \cite{Elsner2001EffectControl,Verschoor2017Self-by-doing:Self-acquisition}.
But how can we model these bidirectional body-schematic action-effect associations computationally, such that an agent can distinguish between its own body, the peripersonal space and the external world? And how does this affect the learning and problem-solving performance of artificial agents?

In this work, we argue that artificial agents/robots require two vital elements to develop the body-schema and peripersonal space representations: (1)~a multimodal sensory integration model, and (2)~a mechanism to learn bidirectional associations between actions and effects. 
The core hypotheses of this paper are that (i) intrinsic rewards created by predictability of an agent's self-caused multimodal sensory effects enable the learning of bidirectional action-effect associations, (ii) these associations implement body schema and peripersonal space representations that make the robotic action-selection robust to noise, and (iii) the abstract body-schematic representations improve the transfer learning performance. 

We address our hypotheses by integrating multisensory association with a forward and inverse mode and a reinforcement learning policy.
Our proposed framework 
(Fig.~\ref{fig:architecture}) is based on the bidirectional associative model by \citet{Pathak2017_curiosity_selfprediction}. 
The model combines the learning of an inverse model and a forward model to emulate the bidirectional action-effect associations. The key property of the model is that it automatically eliminates information from the sensor signals over which the agent has no control, including noise and non-deterministic effects. This is achieved by training simultaneously an abstraction function $\phi(\cdot)$, a forward model $f(\cdot)$ and an inverse model $g(\cdot)$, such that the forward and inverse model operate in the learned abstract state. The abstraction function learns to ignore sensory information that are not directly related to the agent's own actions because the forward and inverse model operate in the abstract space, ignoring all sensorimotor contigencies that are not self-caused \cite{Pathak2017_curiosity_selfprediction}. 
Therefore, the model implicitly learns to distinguish between the agent's own body and the external world. Moreover, the model implies the representations of the agent's body schema and its peripersonal dynamics. 
We exploit this property here to investigate: (1) whether the learned representations benefit from multimodal sensor signals, (2) to what extent they are invariant to noise and (3) whether they are transferable to learn a new skill.

For this investigation, we extend \citeauthor*{Pathak2017_curiosity_selfprediction}'s approach as follows: 
(i) We use multimodal sensory input instead of only visual input for our model; 
(ii) we build on intrinsic motivation to minimize the prediction error, especially in noisy conditions; 
and (iii) we investigate the transfer learning performance. 
In addition, we extend the discrete action space used by  \citeauthor*{Pathak2017_curiosity_selfprediction} towards a continuous action space. 
To realize these extensions, we build on several mechanisms and concepts related to multisensory representation learning, forward models and intrinsic motivation. 




\section{BACKGROUND AND RELATED WORK}
Our proposed model in this paper relates to a body of literature in following main topics:
 
\textbf{Multisensory representation learning of the body schema}
enables humans to perform pose estimation of their body parts, and coordinate transformation between sensory sources, which, ultimately, enables action~\cite{hoffmann_body_2010,Gallese2010TheAction}. Robotic and computational models of body-schematic representations mostly focus on exploiting sensory information from proprioceptive and tactile sensing~\cite{Roncone2014AutomaticRobot, Vicente2016OnlineFeedback, Hoffmann2018RoboticCortex}, or proprioceptive and visual sensing~\cite{Schillaci2014OnlineModel,Wijesinghe2018RobotIntegration, Nguyen2018TransferringTasks} and cast the representation learning as calibration, pose estimation or visuomotor mapping. 

\textbf{The peripersonal space representation} serves as an  interface between the agent's body and the environment \cite{clery_neuronal_2015}. 
Existing robotic and computational models construct the PPS representation from sensory data including vision, audio,  touch and proprioception~\cite{magosso_visuotactile_2010,serino_extending_2015,straka_learning_2017,Chinellato2011ImplicitReaching,juett_learning_2018,Roncone2016PeripersonalSkin,Nguyen2018CompactInteraction,Nguyen_2019_reaching,Pugach2019Brain-inspiredEvents}.
%
Most of the approaches base on the random movements of joints inspired by infants' motor babbling to generate the training data.


\textbf{Forward models} 
are computational models that map the current state of the system to the next state through actions. Some approaches utilize this forward model to learn the imitating actions from multisensory input~\cite{Zambelli:2020:10.1016/j.robot.2019.103312,Copete2017MotorLearning,Hwang2020DealingFramework,Lang2018ARobots}; others employ the forward model to learn the single sensory embedding for control, e.g.~\cite{Watter2015EmbedImagesb,Byravan2018SE3-Pose-Nets:Control,Agrawal2016,Park2018LearningImitate,Pathak2017_curiosity_selfprediction}. Differently, in some neurorobotics models, forward model plays the core role in high-level cognitive functions such as self/other distinction, sense of agency or body ownership~\cite{Lanillos2017YieldingContingencies,Hinz2018DriftingRobot,Lanillos2020RobotMirror}. 



\textbf{Intrinsic motivation} 
is an internal system that drives human to ``engage in particular activities for their own sake, rather than a step towards solving practical problem''~\cite{Oudeyer_2007_IntrinsicMotivation}. Some recent models revisit this concept with the computational models of neural networks (cf.~\cite{Burda2019_ICLR} for an overview).
For example, \citet{Dilokthanakul2019FeatureLearning} implement the intrinsic motivation as the changes in the image features of two consecutive frames, allowing gaming agents to learn by maximizing the change of the visual representations. 
\citet{Pathak2017_curiosity_selfprediction} propose to use prediction error of an additional forward model as intrinsic motivation to drive gaming agents to explore the space. \citet{pathak19disagreement} further extend the idea by using an ensemble of forward functions and exploiting the disagreement among prediction errors in the ensemble as the intrinsic motivation. 
\citet{Roder2020} adopt the ideas of the intrinsic motivation from~\cite{Pathak2017_curiosity_selfprediction} but with only proprioceptive input instead of visual input. 



\section{METHODOLOGY}
\label{sec:method}


\begin{figure}[t]
   \centering
   \includegraphics[width=0.9\linewidth]{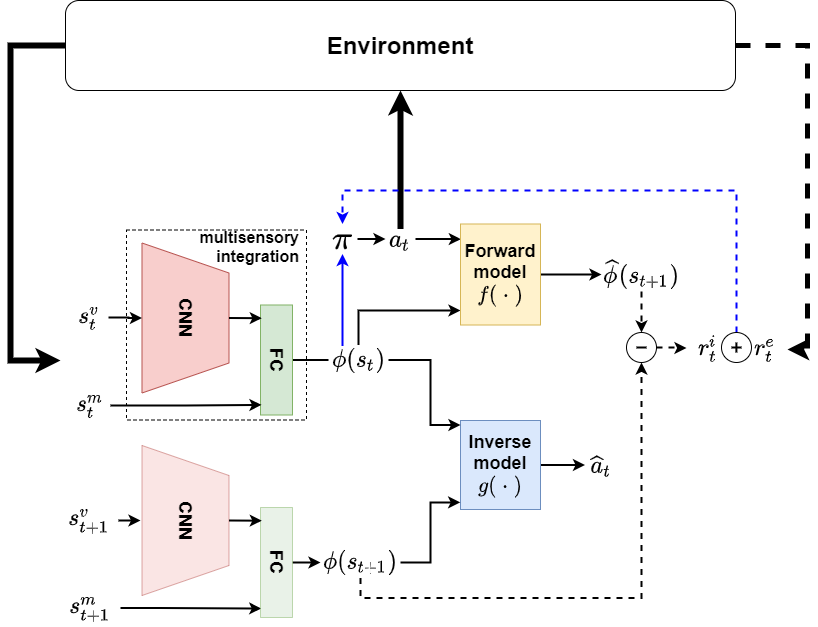}
   \caption{Overview of the learning framework.} 
   \vspace{-0.5cm}
   \label{fig:architecture}
\end{figure}

The objective of our research is to enable a robot to learn to solve a given task in a general goal-directed way, by generalizing the learnt knowledge to any goal \textit{g} in the set of possible goals $\mathcal{G}$. For example, considering the toy example of a planar reaching task (see Fig.~\ref{fig:2j_env}), we aim for a framework that enables a 2DoF-robot to learn to move its end-effector to every position in the plane within the robot's reachable region. More advanced applications involve self-locomotion and object manipulation, where an agent should be able to generalize over goal locations of itself or of external objects.




\subsection{Overview of the framework}
\label{sec:framework}
To achieve the desired generalization and transfer learning abilities, we train a reinforcement learning policy in an abstract generalized state representation imposed by the agent's body schema and peripersonal space model. 
This model is learned with the bio-inspired learning framework presented in Fig.~\ref{fig:architecture}.
The forward model $f(\cdot)$ predicts a sensory effect $\hat{\phi}(s_{t+1})$ from a currently conducted action $a_t$ and the currently perceived sensory state representation $\phi(s_t)$.   
The \textit{policy} $\pi$ generates motor actions $a_t$ under constraints exerted by the environment and under consideration of the prediction error $e_{t+1}$ of the forward model. 
Both the forward model and the policy operate in the latent space of the multimodal sensory input, which is compressed by the \textit{multisensory integration} process. 
We specify the operation of these modules as follows: 
    \begin{align}
        & \text{Multisensory representations:} \notag \\ 
        & \qquad \qquad \phi(s_t) = \phi^{PPS}(s^{e\cup i}_t) \bigcup \phi^{body}(s^i_t) \\
        &\text{Forward model:} \notag \\ 
        & \qquad \qquad \hat{\phi}(s_{t+1}) = f\big( \phi(s_t), a_t;\bm{\theta}_F \big) \\
        & \text{Prediction error:} \notag \\ 
        &\qquad \qquad e_{t+1} = \frac{1}{2} \norm{\hat{\phi}(s_{t+1}) - \phi(s_{t+1})}_2^2 
    \end{align}
where $\phi^{PPS}(s^{e\cup i}_t)$ denotes the representation of the PPS, 
$\phi^{body}(s^i_t)$ denotes the body schema representation, and $\bm{\theta}_F$ denotes parameters of a two-layer fully-connected neural network approximating the function $f(\cdot)$. 

Instead of using the prediction error as an intrinsic motivation to drive agents in seeking for the new information, which is suitable for the navigation task~\cite{Pathak2017_curiosity_selfprediction}, we propose to use the prediction error for a different purpose in this work. For our agents, the lower the prediction error means the better their ability to anticipate their own action, which is represented by the intrinsic reward $r^i_t$ as follows:
\begin{equation}
    r^i_t = -\frac{\lambda}{2}\norm{\hat{\phi}(s_{t+1}) - \phi(s_{t+1})}_2^2    
\end{equation}
where $\lambda$ is a hyperparameter to decide the importance of the intrinsic reward within the whole reward the agents receive. 

In the following sections \ref{sec:multisensory} and \ref{sec:policy} we describe how the intrinsic reward is combined with the extrinsic reward to train an actor-critic reinforcement learning method (cf. Eq.~\ref{eq:q-learning}). Herein, all modules learn simultaneously through the  agent's interactive experience in the environment. 

\subsection{Multisensory integration with neural network}
\label{sec:multisensory}
We aim to enable robots to exploit the relevant information from different sources of their sensory input, to construct a representation of the environment and finally to facilitate the task learning.
Therefore, we first implement a neural network for visuo-motor integration based on \citet{Nguyen2018TransferringTasks}, and construct our input preprocessing network with two branches, one for vision and one for proprioception. 
The former branch consists of four convolution layers with \textit{Elu} activation and a fully-connected layer, while the latter branch contains one layer of fully-connected units. The visual and proprioceptive features are concatenated and combined by another fully-connected layer to produce a compressed latent feature $\phi(s_t)$ of the environment (including the robot itself). The construction of multisensory network is presented in the left side of our general architecture in Fig.~\ref{fig:architecture}. 

Unlike~\citet{Nguyen2018TransferringTasks} and other methods that learn this representation separately, e.g.~\citet{Watter2015EmbedImagesb},~\citet{Zambelli:2020:10.1016/j.robot.2019.103312}, we adopt the approach by~\citet{Pathak2017_curiosity_selfprediction} for state representation learning within the reinforcement framework. While~\citet{Pathak2017_curiosity_selfprediction} consider only visual input, we extend it to multisensory input. In addition, we also employ the inverse model $g(\cdot)$ for representation learning from multiple inputs by minimizing the inverse loss as follows:
\begin{equation}
    \mathcal{L}_I \big(\hat{a}_t, a_t \big) = \frac{1}{2} \norm{\hat{a}_{t} - a_{t}}_2^2, 
\end{equation}
where $\hat{a}_t = g\big(s_t, s_{t+1}; \bm{\theta}_I \big)$ is the estimated action of $a_t$ through the inverse mapping function $g(\cdot)$, approximated by a two-layer fully-connected neural network with parameters $\bm{\theta}_I$. By minimizing the difference between the estimated and real action, the learnt inverse model plays as 
an encoder of the relevant information for the task from multiple sensory input~\cite{Pathak2017_curiosity_selfprediction}. 

\subsection{Reinforcement learning policy}
\label{sec:policy}

We consider a reinforcement learning setting, where an agent interacts with an environment and tries to maximize the long-term expected reward. The interacting environment can be defined as a set of state (or observation) $\mathcal{S}$, an action set $\mathcal{A}$, a reward function $r:\mathcal{S} \times \mathcal{A} \mapsto \mathbb{R}$, transition probability $\rho(s_{t+1}|s_t,a_t):\mathcal{S} \times \mathcal{S} \times \mathcal{A}$,
and a discount rate $\gamma \in [0,1]$. 

In our setting, we employ the policy gradient approach that allows agent to select actions directly through a parameterized policy instead of consulting the value function. This means, at time $t$, the agent takes an action $a_t$ drawn from the   policy $\pi(a|s, \bm{\theta})$, a probability distribution over the action space given the current state $s_t$ with parameter vector $\bm{\theta}$. 

Any policy gradient-based reinforcement learning algorithm can be applied to construct the \textit{policy} in our framework. 
In this work, we employ and validate the Deep deterministic policy gradient algorithm (DDPG)~\cite{Lillicrap2015ContinuousLearning} with continuous actions in combination with our BidAL method. The previous implementation by \citet{Pathak2017_curiosity_selfprediction} uses AC3~\cite{Mnih2016AsynchronousLearningb} with a discrete action space. Moreover, we do not manually design any task-specific rewards, but employ the spare-reward scheme for the external reward $r^e_t$ in our proposal. This means the robots receive 0 reward for successfully completing the desired task and -1 for failing the task, as defined in Eq.~\ref{eq:ext_reward}:
\begin{equation}
    r^e_t = -\big[|g - s_t| > \epsilon\big]
\label{eq:ext_reward}
\end{equation}
where $\epsilon$ is a reasonable threshold value to determine whether an achieved state $s_t$ is close enough to an desired goal $g$ to consider the goal achieved. 
We implement our goal-directed method using deep deterministic policy gradient (DDPG),
universal value function approximators (UVFA)\cite{Schaul2015_UVFA} and hindsight experience replay (HER)\cite{Andrychowicz2017a}, as described below: 

\subsubsection{Deep deterministic policy gradient (DDPG)}
\label{sec:DDPG}
While policy gradient methods in general refer to a parameterized, stochastic policy, deterministic policy gradient methods aim to learn parameters
for a deterministic policy, $\mu_{\bm{\theta}}:\mathcal{S} \mapsto \mathcal{A}$. 
\citet{Silver2014DeterministicAlgorithms} shows that the deterministic policy gradient (DPG) exists as a special case of stochastic policy, but can be estimated more efficiently. 
The efficient exploration of a deterministic policy can be guaranteed with an off-policy algorithm 
in which actions are chosen according to a stochastic behaviour policy $\beta(a|s)$, but to learn about a deterministic target policy~\cite{Silver2014DeterministicAlgorithms}. The off-policy deterministic policy gradient is written as:
\begin{equation}
    \begin{split}
        \nabla_{\bm{\theta}}J_\beta(\mu_{\bm{\theta}}) 
        & \approx \int_\mathcal{S} \rho^\beta(s) \nabla_{\bm{\theta}}\mu_{\bm{\theta}} Q^{\mu}(s,a) ds\\   
        & = \E_{s\sim \rho^\beta} \Big[ \nabla_{\bm{\theta}}\mu_{\bm{\theta}}(s) \nabla_a Q^\mu(s,a)|_{a=\mu_{\bm{\theta}}(s)} \Big]
    \end{split}
    \label{eq:off-policy-gradient}
\end{equation}
where $\rho^\beta$ denotes the state distribution of $\beta(a|s)$.

DDPG~\cite{Lillicrap2015ContinuousLearning} extends the actor-critic approach of DPG with neural network function approximators. The actor network of 
$\mu_{\bm{\theta}}$ is updated with deterministic policy gradient as Eq.~\ref{eq:off-policy-gradient}, while the critic network estimates the action-value function $Q(s,a)$ by minimizing the loss in Eq.~\ref{eq:q-learning}. 
\begin{equation}
\begin{split}
    & \mathcal{L} = \E_{s_t\sim \rho^\beta, a_t \sim \beta} \Big[\big( Q(s_t,a_t|\bm{\theta}^Q) - y_t \big)^2 \Big]\\
    \text{with target: }
    & y_t = r_t + \gamma Q(s_{t+1}, \mu{s_{t+1}| \bm{\theta}^Q})\\
    \text{and reward: }
    & r_t = r^i_t + r^e_t
\end{split}
    \label{eq:q-learning}
\end{equation}
Both actor and critic neural networks are updated by sampling from a finite sized replay buffer where tuples of exploration $(s_t, a_t, r_t, s_{t+1})$ have been stored. 

\subsubsection{Universal value function approximation (UVFA)}
In order to generalize the learnt policy to both the state space and the goal space,~\cite{Schaul2015_UVFA} proposes the UVFA approach to represent a set of optimal action-value functions by using a unified function approximator. This is achieved by extending the Q-function to depend not only on the state-action $(s,a)$ pair but also the goal $g$. In terms of function approximation with neural networks, we can concatenate $(s,g)$ or their embeddings $(\psi(s),\eta(g))$  for the actor network, and $(s,a,g)$ or $(\psi(s),a,\eta(g))$ for the critic network. Our networks follow the embeddings structure. This technique is important for our setup 
as it allow robots to learn a general goal-directed policy: Instead of achieving a specific task, e.g. reaching a certain position, they learn to complete a more general one, i.e. reaching every position within the reachable space. This also makes our approach different from the work by \citet{Pathak2017_curiosity_selfprediction}.


\subsubsection{Hindsight experience replay (HER)} We employ this technique from~\citet{Andrychowicz2017a} to enrich the replay buffer by assuming that some random unsuccessful achieved state is the goal state. 
\subsection{Evaluation environments}
\label{sec:env}
\begin{figure}[!th]
     \centering
     \begin{subfigure}[b]{0.4\linewidth}
         \centering
          \includegraphics[width=\linewidth]{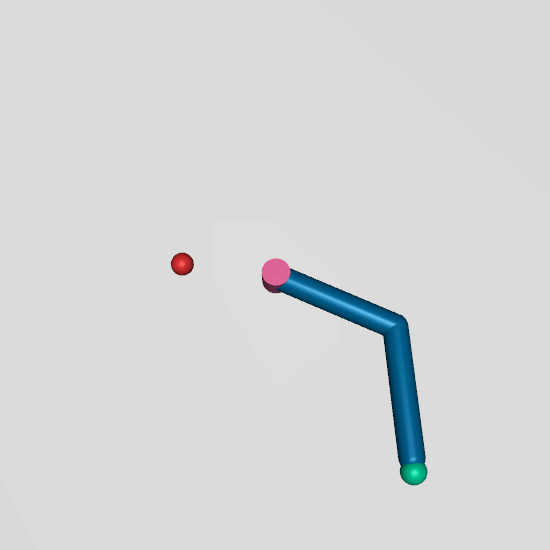}
         \caption{Planar reaching}
          \label{fig:2j_env}
     \end{subfigure}
     \begin{subfigure}[b]{0.4\linewidth}
         \centering
         \includegraphics[width=\linewidth]{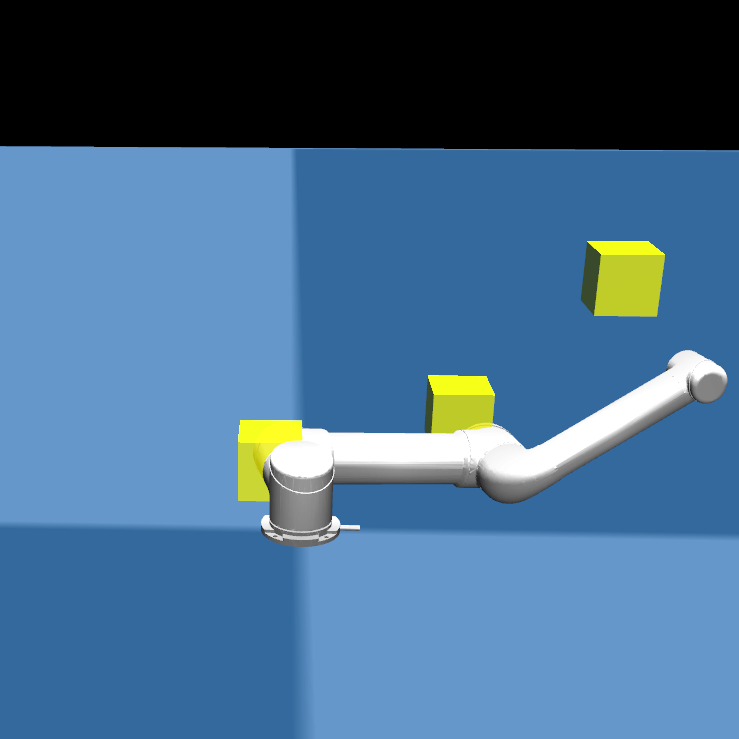}
         \caption{UR5 reaching}
         \label{fig:ur5_env}
     \end{subfigure}
     \vspace{-0.1cm}
     \begin{subfigure}[b]{0.4\linewidth}
         \centering
          \includegraphics[width=\linewidth]{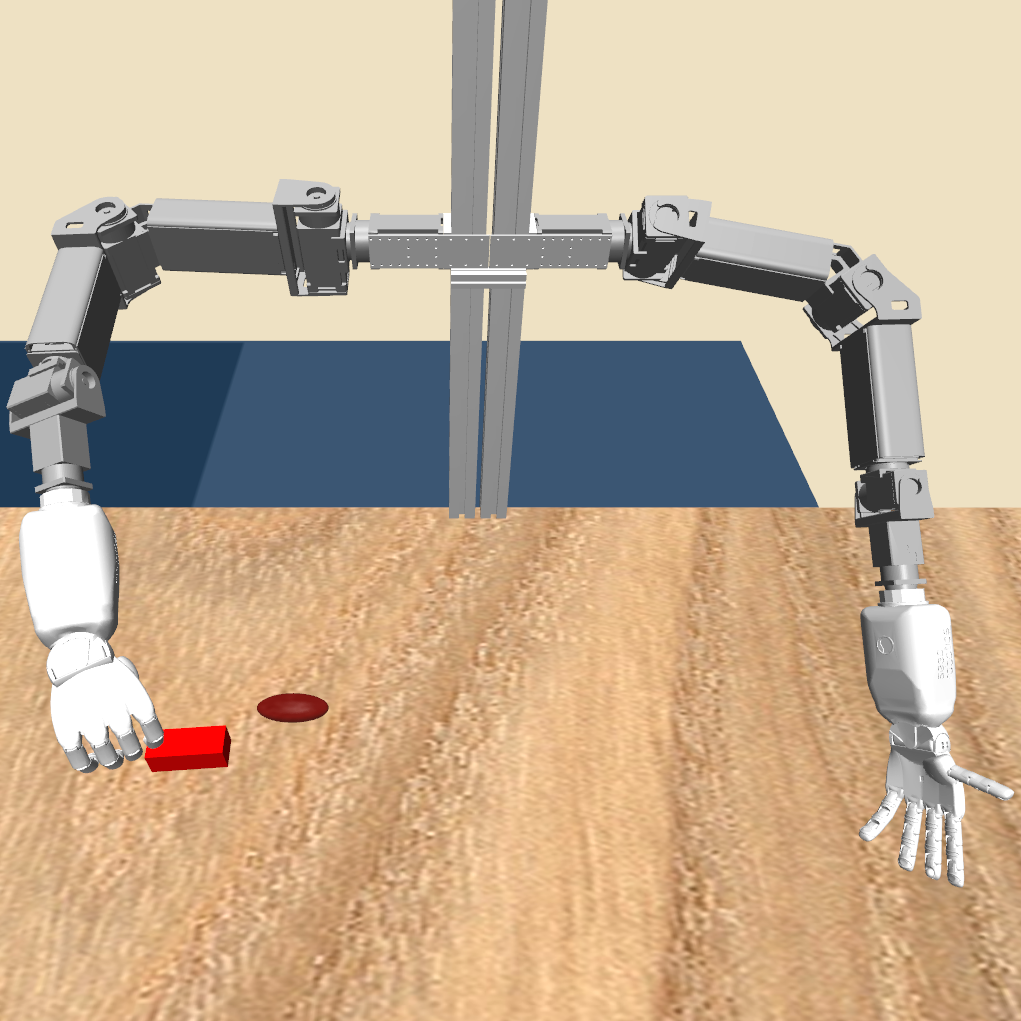}
         \caption{Object sliding}
          \label{fig:nicol_slide_env}
     \end{subfigure}
     \begin{subfigure}[b]{0.4\linewidth}
         \centering
          \includegraphics[width=\linewidth]{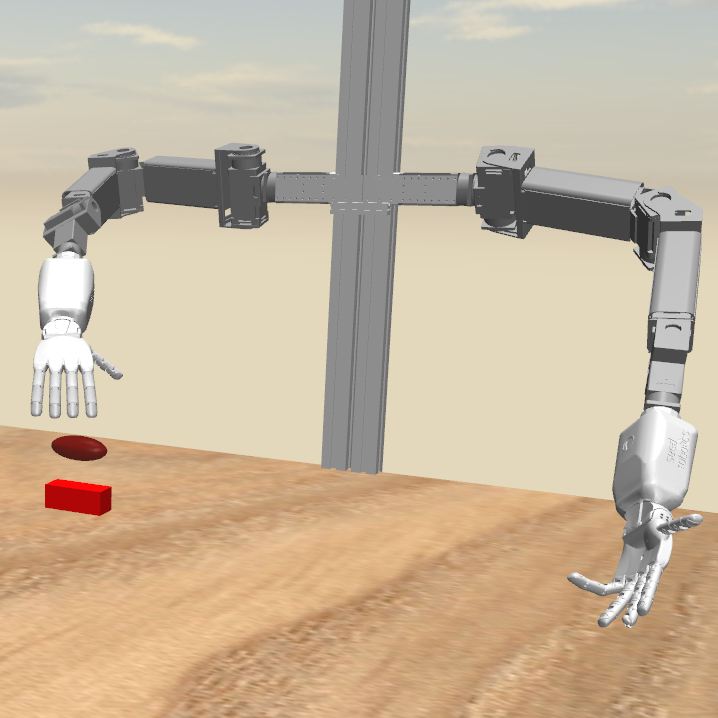}
         \caption{Object lifting}
          \label{fig:nicol_lift_env}
     \end{subfigure}
     \caption{Experimental environments}
     \vspace{-0.45cm}
     \label{fig:env}
\end{figure}

\begin{table}[!h]
\centering
\begin{tabular}{M{.1\textwidth} | M{.1\textwidth} | M{.12\textwidth} | M{.06\textwidth}} 
\hline
Env. & $\mathbf{s}^v_t$ & $\mathbf{s}^m_t$ & $\mathbf{a}_t$ \\
\hline
Planar reaching & \multirow{4}{*}{$\Big[\mathbf{I}\Big]_{(42,42,1)}$} & $\Big[\mathbf{q},\mathbf{\dot{q}}\Big]_{(0,4)}$ & [0,2] \\
\cline{1-1} \cline{3-4}
UR5 reaching &  & $\Big[\mathbf{q},\mathbf{\dot{q}}\Big]_{(0,6)}$ & [0,3] \\
\cline{1-1} \cline{3-4}
Object sliding &  & \multirow{2}{*}{$\Big[\mathbf{x},\mathbf{q},\mathbf{\dot{x}},\mathbf{\dot{q}}\Big]_{(0,40)}$} & \multirow{2}{*}{[0,7]*} \\
\cline{1-1} 
Object lifting &  &  &  \\
\hline
\end{tabular}
\caption{Summary of experimental environments. For object manipulation tasks$^{(*)}$, i.e. sliding and lifting, the agent can control the position of the arm in the Cartesian space, roll and tilt its hand, and move all fingers and its thumb abduction.}
\label{tab:env}
\vspace{-.6cm}
\end{table}

We evaluate our method by conducting experiments in simulated environments based on the Mujoco software~\cite{Todorov2012MuJoCo:Control}. Depending on the specific environment, the setup can be varied but generally all robots have the common properties:

\begin{figure*}[!h]
     \centering
     \begin{subfigure}[b]{0.35\linewidth}
         \centering
          \includegraphics[width=\linewidth]{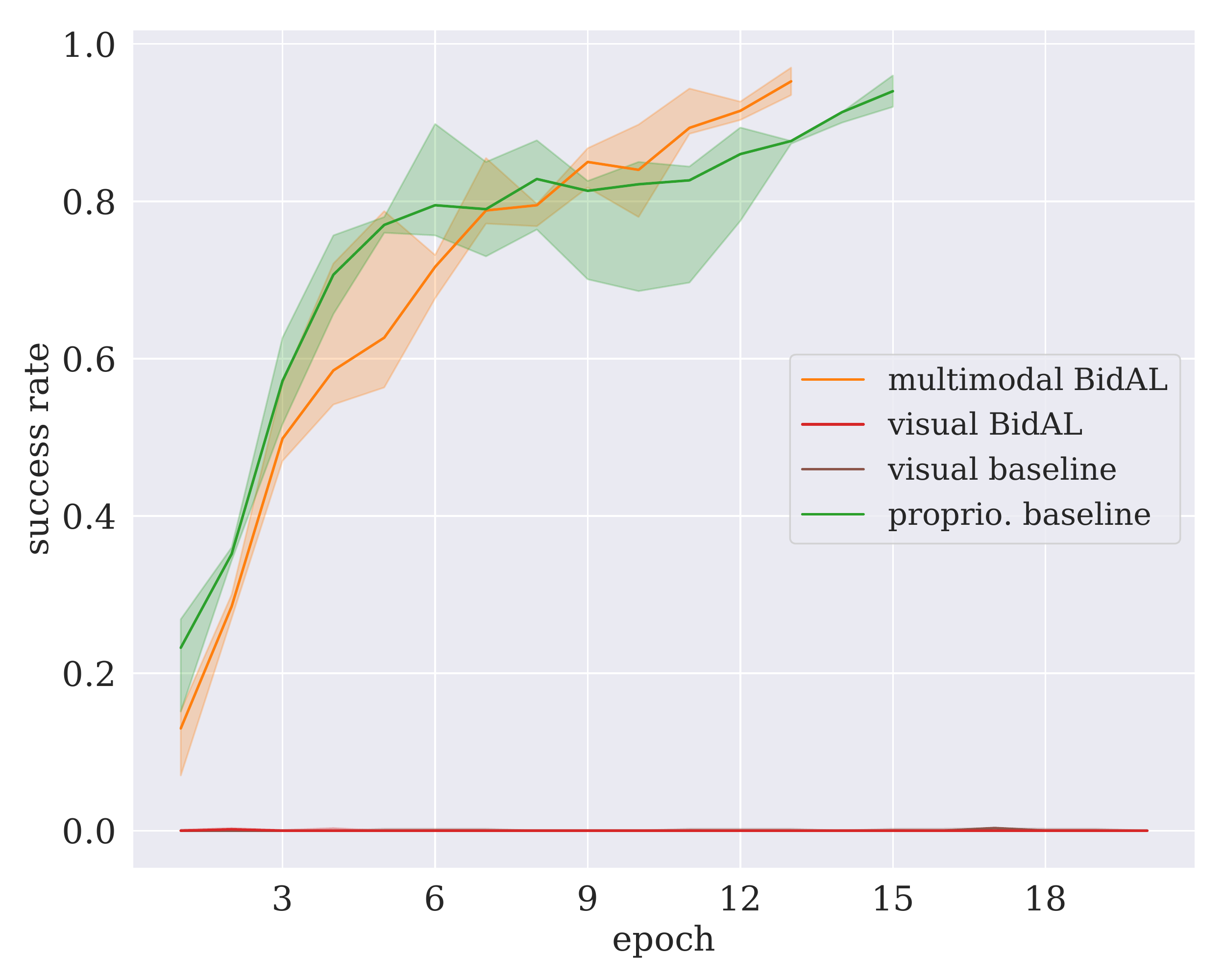}
         \caption{Planar reaching}
          \label{fig:2j_learn}
     \end{subfigure}
     \begin{subfigure}[b]{0.35\linewidth}
         \centering
         \includegraphics[width=\linewidth]{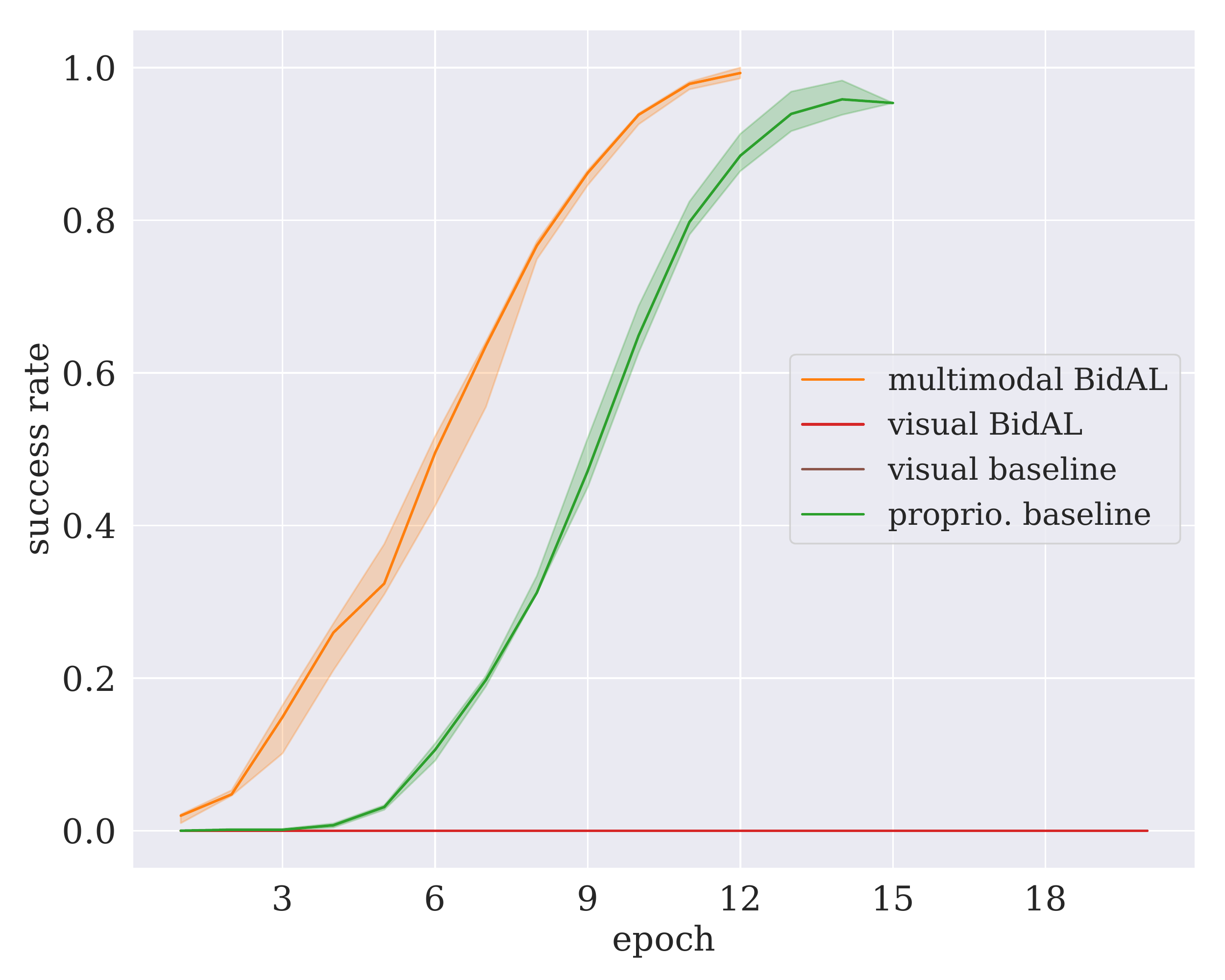}
         \caption{UR5 reaching}
         \label{fig:ur5_learn}
     \end{subfigure}
     \begin{subfigure}[b]{0.35\linewidth}
         \centering
          \includegraphics[width=\linewidth]{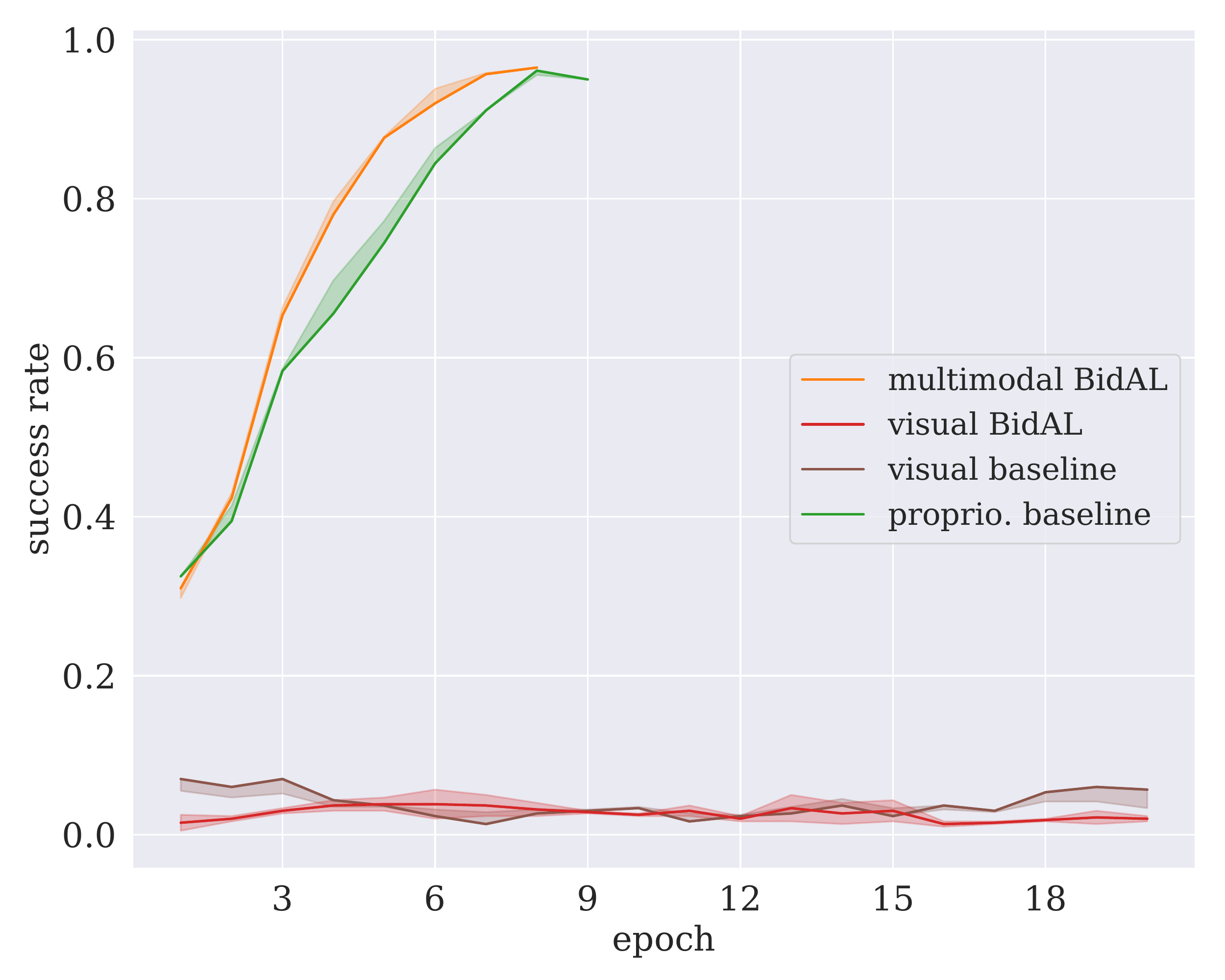}
         \caption{Object sliding}
          \label{fig:nicol_slide_learn}
     \end{subfigure}
     \begin{subfigure}[b]{0.35\linewidth}
         \centering
          \includegraphics[width=\linewidth]{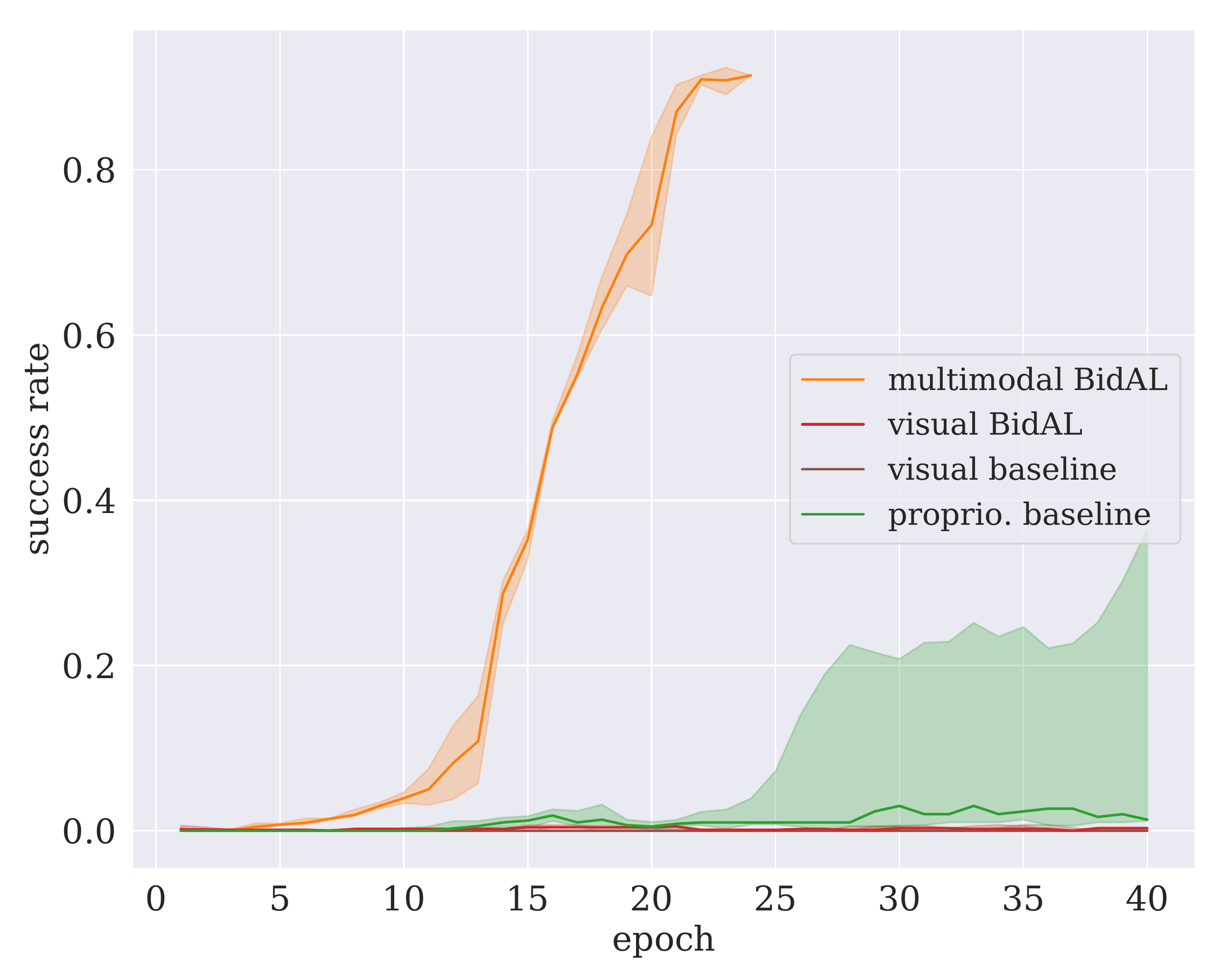}
         \caption{Object lifting}
          \label{fig:nicol_slide_lift}
     \end{subfigure}
     \caption{Learning performance in different environment. Note that we use the term ``proprioception'' for both joint and Cartesian measurements. See Section~\ref{sec:env} for more details}
     \label{fig:exp_learn}
\end{figure*}  

\begin{itemize}
    \item They have access to cameras which provide rendered RGB frames of the environment, including parts of the robot itself. The rendered frames are then converted to gray format and scaled down to smaller dimension. 
    This source of input is denoted as $\mathbf{s}^v_t$ in this work;
    \item Robots have access to encoder measurements of all joints composed of their body. Additionally, the end-effector and object information are available for object manipulation environment. We use the term ``proprioception'' for this input in an extended meaning, and denote it as $\mathbf{s}^m_t$ in this work; 
    \item Agents generate action in only the joint space (in reaching tasks) or mixes of the joint and the Cartesian space (in object manipulation environments). The size of generated actions also varies in different environments.
\end{itemize}


We illustrate the specific environments that we use for our experiments in Fig.~\ref{fig:env}:
\subsubsection{Planar reaching} is a robot with two revolute joints and the end-effector depicted with the green dot. Its task is to reach a target on a 2D plane (depicted with the red dot) without a priori knowledge of its kinematics model.

\subsubsection{UR5 reaching} is composed of the three first joints of a UR5 robot\footnote{\url{https://www.universal-robots.com/de/produkte/ur5-roboter/}}. The goal of the robot in this environment is to explore the 3D space and reach the desired joint angles (shown by the three yellow cubes).

\subsubsection{NICOL manipulator} is a humanoid torso composed of two OpenManipulator-P arms\footnote{\url{http://www.robotis.us/openmanipulator-p/}}  and two R8H hands\footnote{\url{https://www.seedrobotics.com/rh8d-adult-robot-hand}}. The robot is tasked to learn in varieties of object manipulating scenarios: (i) Reach and move a cube to a desired position (marked by the dark red ellipsoid) on the table--object sliding, or (ii) reach and lift an object up to the air--object lifting. Note that the lifting task implicitly requires the robot to learn a robust grasping skill to complete.

\section{EXPERIMENTS \& RESULTS}
\label{sec:exps-result}
We evaluate our proposed method in three experiments. We first train the agent with combined intrinsic and extrinsic rewards, then we investigate the robustness to noise, and, finally, we investigate the transfer learning performance.
\begin{figure*}[!t]
     \centering
     \begin{subfigure}[b]{0.325\linewidth}
         \centering
          \includegraphics[width=\linewidth]{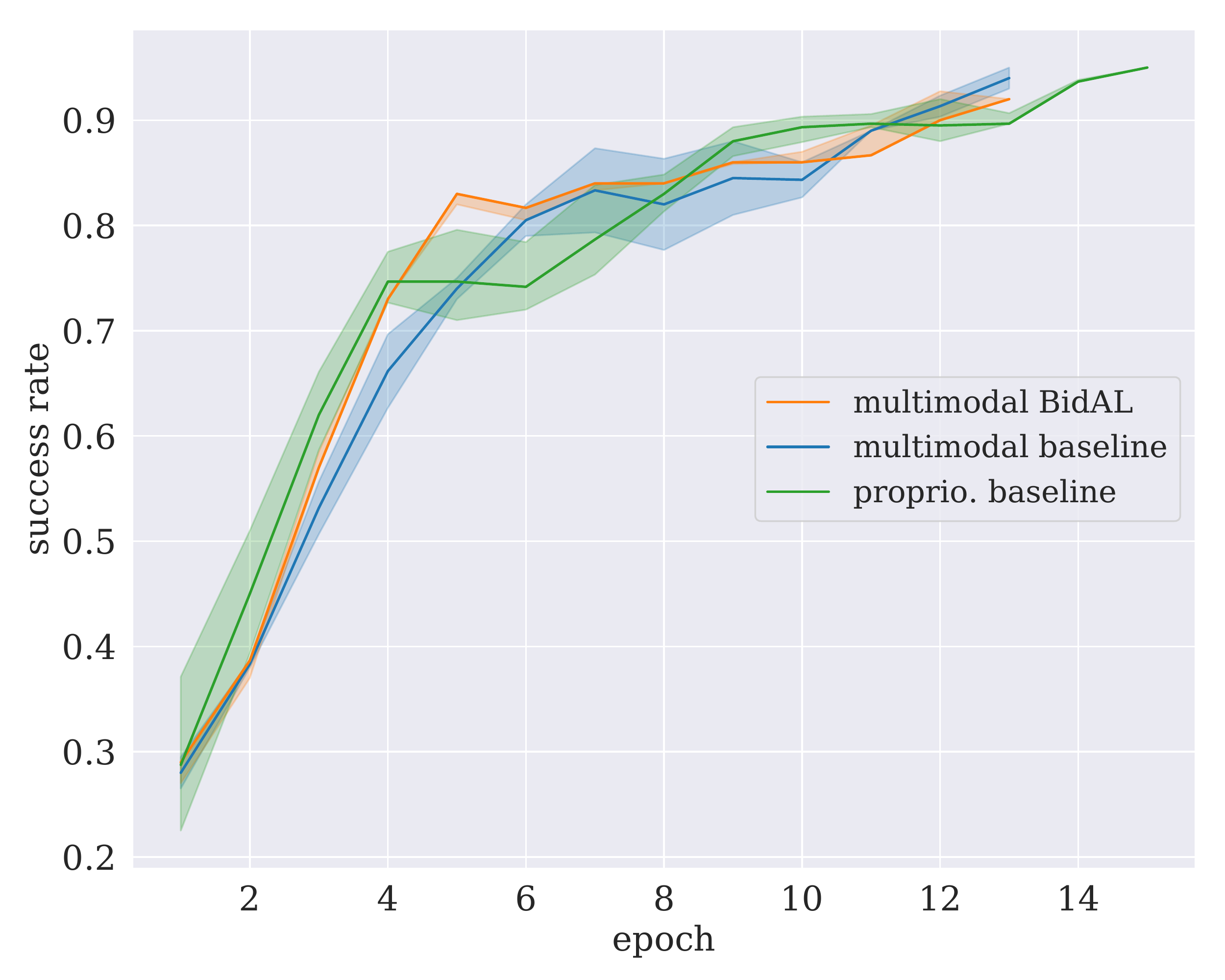}
     \end{subfigure}
     \begin{subfigure}[b]{0.325\linewidth}
         \centering
         \includegraphics[width=\linewidth]{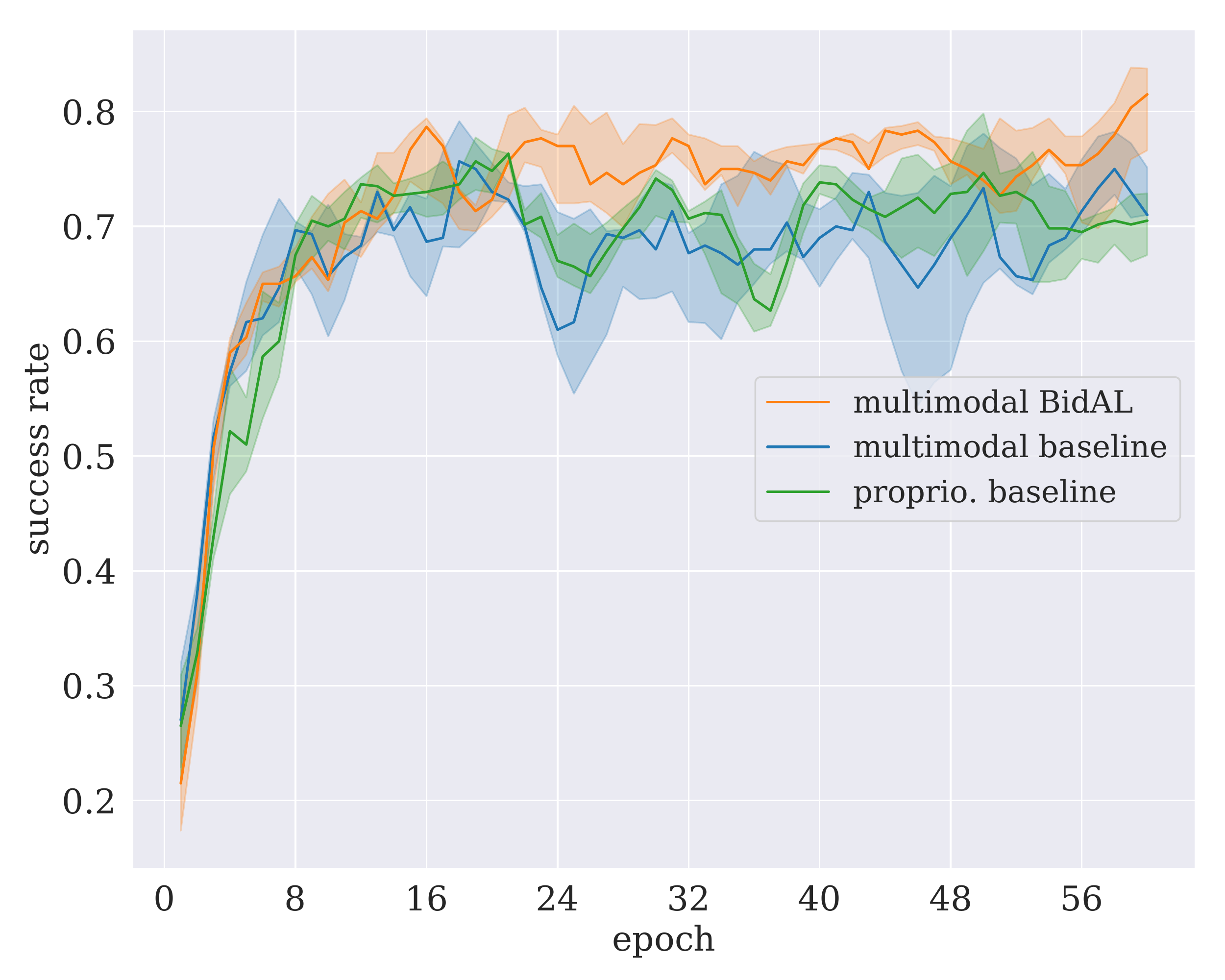}
     \end{subfigure}
     \begin{subfigure}[b]{0.325\linewidth}
         \centering
          \includegraphics[width=\linewidth]{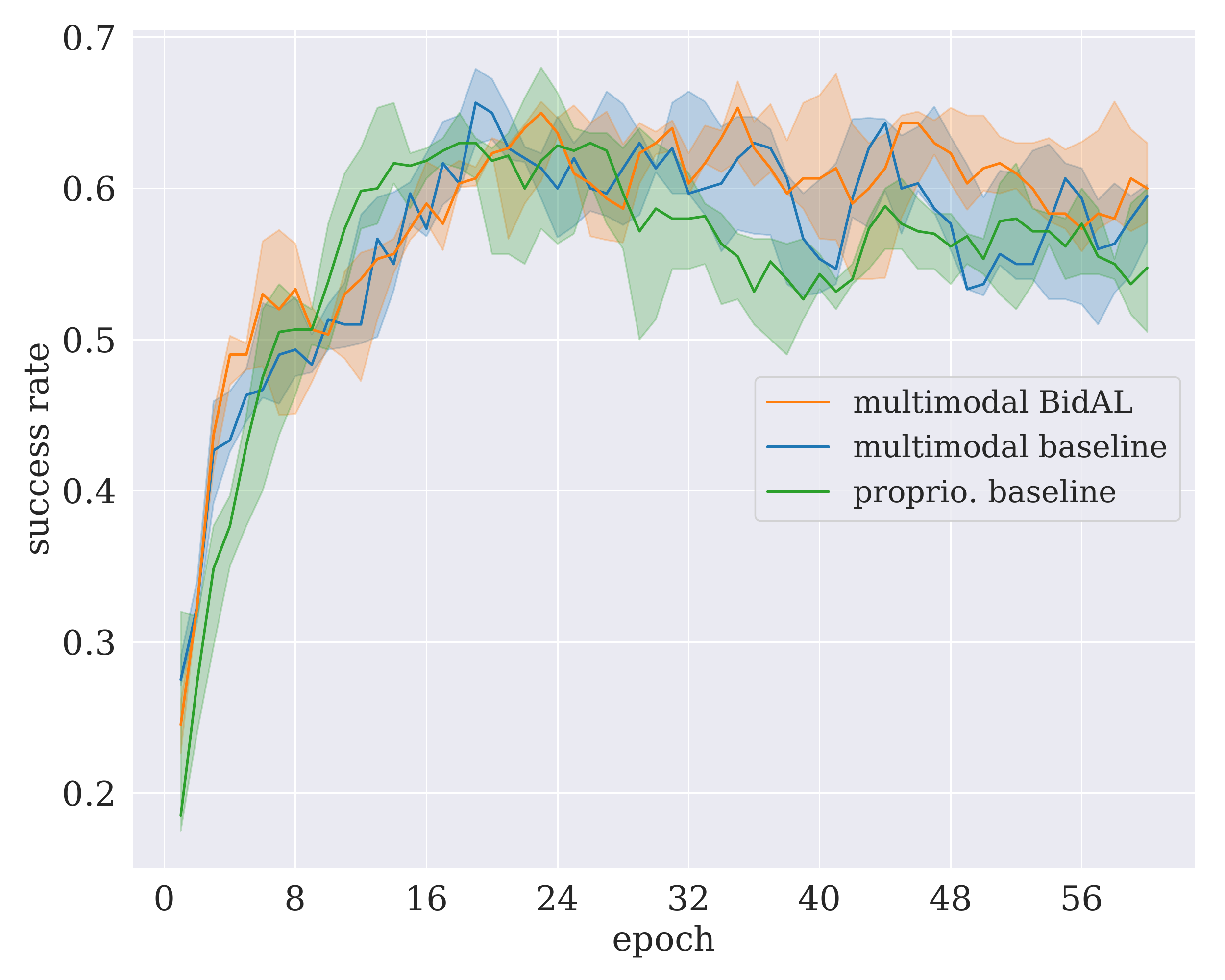}
     \end{subfigure}
     \caption{Learning performance in different noise conditions for the object sliding tasks: 5\%--Left, 10\%--Center and 15\%--Right}
     \vspace{-0.5cm}
     \label{fig:exp_noise}
\end{figure*}  
\subsection{Learning performance with combined reward}
\label{sec:exp1}

To investigate how combined intrinsic and extrinsic rewards affect the learning performance, the robot  receives two types of reward for its  actions: A sparse extrinsic reward from the environment for achieving the task and an intrinsic reward computed from the prediction error.

Herein, we investigate four configurations:
As baselines, we use DDPG+HER with proprioceptive input (proprio. baseline) and visual input (visual baseline). 
We compare these baselines with our proposed algorithm, where we combine the bidirectional associative learning (BidAL) method with the visual input (visual BidAL) and multisensory integration (multimodal BidAL).
For these configurations, we perform training in all simulated environments presented in Section~\ref{sec:env}. 

The results in Fig.~\ref{fig:exp_learn} show that our proposed algorithm learns all tasks efficiently, while the baselines with only visual input fail to learn any task.
The baselines with proprioceptive input perform significantly better than those with visual input. 
Overall, our approach outperforms all baselines, especially in the object lifting task. This result provides direct evidence for our core hypothesis that the bidirectional associations improve the learning performance.
The results suggest further that minimizing the prediction loss is beneficial for learning the multisensory representation, which is a prerequisite for the robots to learn the main desired task. 


\subsection{Robustness to noise}
\label{sec:exp2}
We further investigate the role of proprioception in the multimodal setting by adding observational noise to the end-effector pose and the object position.
This experiment focuses on the task of object sliding (see Fig.~\ref{fig:nicol_slide_env}), and aims to simulate that the robot's end-effector and object pose cannot be measured directly in the real world. Normally, these information is obtained through additional estimation processes, mostly from visual input. These estimations may contribute to noise or inaccuracy in the general observation. We realize the noisy observations as follows:
\begin{equation}
    s_{noisy} = s + \kappa \cdot n
\end{equation}
where $\kappa$ denotes the noise coefficient, $n$ denotes noise and $n \sim \mathcal{N}(0,\,\sigma)$, $\sigma$ denotes the range between $75^{th}$ and $25^{th}$ percentile of the continuous history of $s$.   

We perform the training with the combined intrinsic and extrinsic rewards at three different values of the noise coefficient, namely 5\%, 10\% and 15\%.
Fig.~\ref{fig:exp_noise} illustrates that a high noise coefficient affects the performance significantly in all experiments. 
However, 
our BidAL approach is more robust to noise, performing significantly more stable than the baseline as
we compare the mean and standard deviation of the success rate over the last $N$ epochs\footnote{$N=30$ for 10-15\% noise and $N=5$ for 5\% noise.} over 5 training runs (see \autoref{tab:noise_stddev}).

\begin{table}[!ht]
    \centering
    \vspace{-0.3cm}
    \begin{tabular}{r|c c c}
        Noise ($\kappa$) & multi. BidAL &  multi. baseline&
        proprio. baseline\\
        \hline
        5\%   & $0.874 \pm 0.0127$ & $0.886 \pm 0.0143$ & $0.882 \pm 0.0191$  \\
        10\%  & $0.764 \pm 0.0222$ & $0.696 \pm 0.0542$ & $0.699 \pm 0.0519$  \\
        15\%  & $0.615 \pm 0.0320$ & $0.590 \pm 0.0410$ & $0.554 \pm 0.0358$ \\  
    \end{tabular}
    \caption{Evaluation of robustness to noise.} 
    \vspace{-0.3cm}
    \label{tab:noise_stddev}
\end{table}



For the mean success rates, we see that a drop in the success rate when comparing 5\% noise with 15\% noise: by a factor of 0.70 and 0,66 for the multimodal BidAL and multimodal baseline, respectively. 
The higher factor of the BidAL approach illustrates a slight increase in the robustness to noise. 

To investigate the stability of the learning under noisy conditions we consider the mean standard deviation of the success rate.
With 10\% noise, the mean standard deviation of the multimodal BidAL approach is 0.022, compared to a significantly larger value of 0.054 for the multimodal baseline. Hence, the mean standard deviation with the BidAL approach is less then half (factor of 0.4) of the multimodal baseline. 
In the case of 15\% noise, 
these numbers are 0.032 and 0.041 respectively:
BidAL mean standard deviation is around 3/4 (factor of 0.78) of the baseline.

Furthermore, both metrics favor our multimodal BidAL over the proprioceptive baseline.


\subsection{Transfer learning from a simple to a complex skill}
\label{sec:exp3}

\begin{figure}[t]
    \centering
        \includegraphics[width=0.8\linewidth]{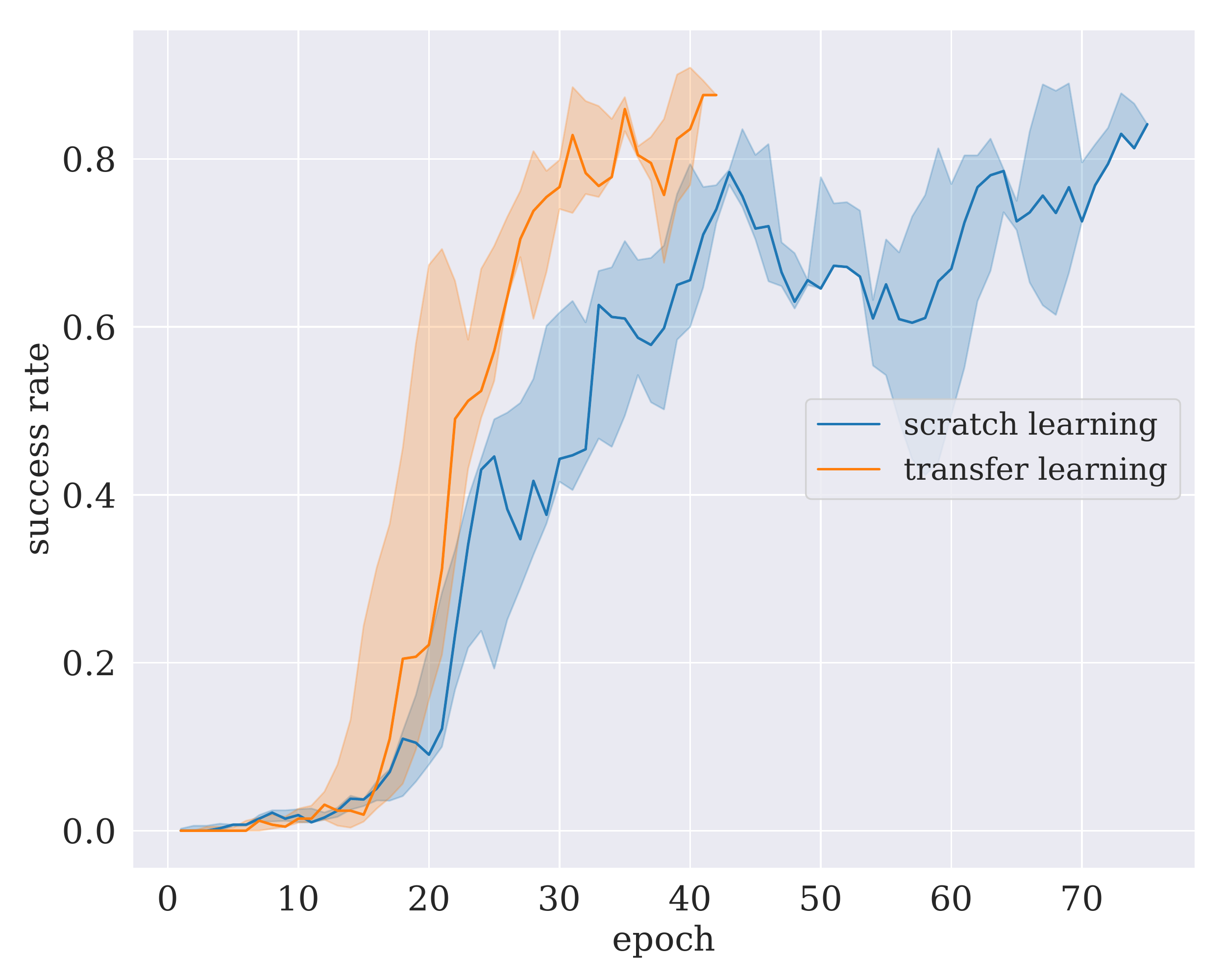}
    \caption{The robot learns the task of object lifting from multimodal input, with and without transfer learning} 
    \vspace{-0.7cm}
    \label{fig:transfer}
\end{figure}
This experiment addresses our hypothesis that previously learned  peripersonal space and body schema representations, encoded in the forward and inverse model, foster the learning of new tasks. 
We propose to pre-train these representations first with a simple skill, and then re-use the representations later to learn a novel skill. 
Therefore, we first train the NICOL robot to slide objects (see Fig.~\ref{fig:nicol_slide_env}), and then continue learning the more complex task of lifting an object (see Fig.~\ref{fig:nicol_lift_env}). 

To obtain more meaningful results, we increase the difficulty of the lifting task by reducing proprioceptive inputs of the robot that are important for grasping. Specifically, we remove the object orientation, the object velocity, and the relative distance between object and end-effector. 
This reduced input explains the poorer performance of lifting task in this section than in Section~\ref{sec:exp1} when the robot learns from scratch. 
The results in Fig.~\ref{fig:transfer}
clearly show that through pre-training the robot with the sliding task, it can more quickly learn to complete the lifting task.



\section{CONCLUSIONS}
\label{sec:conclusion}
We have introduced a new goal-directed reinforcement learning framework that builds on multisensory self-representations, realized as bidirectional action-effect associations.
Our approach aims to allow agents to exploit relevant sensory input and interact with the environment in a continuous manner.
We evaluate our approach in different continuous environments. 
Our three experiments address our three main hypotheses. We show that intrinsic predictability-driven rewards enable the learning  of  bidirectional  action-effect  associations, and we show that these associations implement  body  schema  and  peripersonal  space representations  that  make  the  robotic  action-selection  robust to noise. Finally, we demonstrate that our approach is beneficial for transfer learning.





Future work involves optimization of hyperparamters and the combination with a planning method. 
The planning will be based on the forward model that we have learned.
We hypothesize that integrating planning with a reinforcement learning policy will further improve the learning performance, as demonstrated by related approaches  \cite{Srinivas2018UniversalNetworks,Qureshi2019MotionNetworks,Eppe_2019_semantics}.


\section*{ACKNOWLEDGMENT}
We thank Nicolas Frick for the help on the NICOL design and part of the NICOL simulation used in this paper.
The authors acknowledge funding by the German  Research  Foundation (DFG) through the  IDEAS and LeCAREbot projects.

\printbibliography

@article{magosso_visuotactile_2010,
	title = {Visuotactile representation of peripersonal space: a neural network study},
	volume = {22},
	shorttitle = {Visuotactile representation of peripersonal space},
	url = {http://www.mitpressjournals.org/doi/abs/10.1162/neco.2009.01-08-694},
	number = {1},
	urldate = {2016-05-31},
	journal = {Neural computation},
	author = {Magosso, Elisa and Zavaglia, Melissa and Serino, Andrea and Di Pellegrino, Giuseppe and Ursino, Mauro},
	year = {2010},
	pages = {190--243}
}

@article{clery_neuronal_2015,
	title = {Neuronal bases of peripersonal and extrapersonal spaces, their plasticity and their dynamics: {Knowns} and unknowns},
	volume = {70},
	issn = {00283932},
	shorttitle = {Neuronal bases of peripersonal and extrapersonal spaces, their plasticity and their dynamics},
	url = {http://linkinghub.elsevier.com/retrieve/pii/S0028393214003820},
	doi = {10.1016/j.neuropsychologia.2014.10.022},
	language = {},
	urldate = {2016-05-31},
	journal = {Neuropsychologia},
	author = {Cléry, Justine and Guipponi, Olivier and Wardak, Claire and Ben Hamed, Suliann},
	month = apr,
	year = {2015},
	pages = {313--326}
}

@article{serino_extending_2015,
	title = {Extending peripersonal space representation without tool-use: evidence from a combined behavioral-computational approach},
	volume = {9},
	issn = {1662-5153},
	shorttitle = {Extending peripersonal space representation without tool-use},
	url = {http://journal.frontiersin.org/Article/10.3389/fnbeh.2015.00004/abstract},
	doi = {10.3389/fnbeh.2015.00004},
	urldate = {2016-05-31},
	journal = {Frontiers in Behavioral Neuroscience},
	author = {Serino, Andrea and Canzoneri, Elisa and Marzolla, Marilena and di Pellegrino, Giuseppe and Magosso, Elisa},
	month = feb,
	year = {2015}
}

@article{hoffmann_body_2010,
	title = {Body {Schema} in {Robotics}: {A} {Review}},
	volume = {2},
	issn = {1943-0604, 1943-0612},
	shorttitle = {Body {Schema} in {Robotics}},
	url = {http://ieeexplore.ieee.org/lpdocs/epic03/wrapper.htm?arnumber=5601749},
	doi = {10.1109/TAMD.2010.2086454},
	number = {4},
	urldate = {2016-06-14},
	journal = {IEEE Transactions on Autonomous Mental Development},
	author = {Hoffmann, Matej and Marques, Hugo and Arieta, Alejandro and Sumioka, Hidenobu and Lungarella, Max and Pfeifer, Rolf},
	month = dec,
	year = {2010},
	pages = {304--324}
}

@article{head_sensory_1911,
	title = {{Sensory} {disturbances} {from} {cerebral} {lesions}},
	volume = {34},
	url = {http://brain.oxfordjournals.org/content/34/2-3/102.abstract},
	doi = {10.1093/brain/34.2-3.102},
	number = {2-3},
	journal = {Brain},
	author = {Head, Henry and Holmes, Gordon},
	month = nov,
	year = {1911},
	pages = {102--254}
}

@inproceedings{straka_learning_2017,
	title = {Learning a peripersonal space representation as a visuo-tactile prediction task},
	pages = {101--109},
	booktitle = {International Conference on Artificial Neural Networks},
	publisher = {Springer},
	author = {Straka, Zdenek and Hoffmann, Matej},
	year = {2017}
}

@inproceedings{juett_learning_2018,
	title = {Learning to Grasp by Extending the Peri-Personal Space Graph},
	pages = {8695--8700},
	booktitle = {2018 {IEEE}/{RSJ} International Conference on Intelligent Robots and Systems},
	publisher = {{IEEE}},
	author = {Juett, Jonathan and Kuipers, Benjamin},
	year = {2018},
	langid = {english}
}

@inproceedings{Pathak2017_curiosity_selfprediction,
    title = {{Curiosity-driven Exploration by Self-supervised Prediction}},
    year = {2017},
    booktitle = {International Conference on Machine Learning (ICML)},
    author = {Pathak, Deepak and Agrawal, Pulkit and Efros, Alexei A. and Darrell, Trevor},
    month = {5},
    url = {http://arxiv.org/abs/1705.05363},
    arxivId = {1705.05363}
}

@inproceedings{Roder2020,
    title = {{Curious Hierarchical Actor-Critic Reinforcement Learning}},
    year = {2020},
    booktitle = {29th International Conference on Artificial Neural Networks (ICANN2020)},
    author = {R{\"{o}}der, Frank and Eppe, Manfred and Nguyen, Phuong D. H. and Wermter, Stefan},
    url = {http://arxiv.org/abs/2005.03420},
    arxivId = {2005.03420}
}

@article{Eppe_2019_semantics,
    title = {{From Semantics to Execution: Integrating Action Planning With Reinforcement Learning for Robotic Causal Problem-Solving}},
    year = {2019},
    journal = {Frontiers in Robotics and AI},
    author = {Eppe, Manfred and Nguyen, Phuong D H and Wermter, Stefan},
    pages = {123},
    volume = {6},
    url = {https://www.frontiersin.org/article/10.3389/frobt.2019.00123},
    doi = {10.3389/frobt.2019.00123},
    issn = {2296-9144}
}

@inproceedings{Andrychowicz2017a,
    title = {{Hindsight Experience Replay}},
    year = {2017},
    booktitle = {Conference on Neural Information Processing Systems (NIPS)},
    author = {Andrychowicz, Marcin and Wolski, Filip and Ray, Alex and Schneider, Jonas and Fong, Rachel and Welinder, Peter and McGrew, Bob and Tobin, Josh and Abbeel, Pieter and Zaremba, Wojciech},
    pages = {5048--5058},
    url = {https://papers.nips.cc/paper/7090-hindsight-experience-replay.pdf},
    arxivId = {1707.01495}
}

@article{Oudeyer_2007_IntrinsicMotivation,
    title = {{Intrinsic Motivation Systems for Autonomous Mental Development}},
    year = {2007},
    journal = {IEEE Transactions on Evolutionary Computation},
    author = {Oudeyer, P and Kaplan, F and Hafner, V V},
    number = {2},
    month = {4},
    pages = {265--286},
    volume = {11},
    doi = {10.1109/TEVC.2006.890271}
}

@inproceedings{Burda2019_ICLR,
    title = {{Large-Scale Study of Curiosity-Driven Learning}},
    year = {2019},
    booktitle = {International Conference on Learning Representations (ICLR)},
    author = {Burda, Yuri and Edwards, Harri and Pathak, Deepak and Storkey, Amos and Darrell, Trevor and Efros, Alexei A},
    url = {https://pathak22.github.}
}

@inproceedings{Agrawal2016,
    title = {{Learning to poke by poking: Experiential learning of intuitive physics}},
    year = {2016},
    booktitle = {Advances in Neural Information Processing Systems},
    author = {Agrawal, Pulkit and Nair, Ashvin and Abbeel, Pieter and Malik, Jitendra and Levine, Sergey},
    pages = {5074--5082},
    issn = {10495258},
    arxivId = {1606.07419}
}

@article{Zambelli:2020:10.1016/j.robot.2019.103312,
    title = {{Multimodal representation models for prediction and control from partial information}},
    year = {2020},
    journal = {Robotics and Autonomous Systems},
    author = {Zambelli, M and Cully, A and Demiris, Y},
    volume = {123},
    url = {http://dx.doi.org/10.1016/j.robot.2019.103312},
    doi = {10.1016/j.robot.2019.103312}
}

@inproceedings{Nguyen_2019_reaching,
    title = {{Reaching development through visuo-proprioceptive-tactile integration on a humanoid robot - a deep learning approach}},
    year = {2019},
    booktitle = {ICDL-EpiRob 2019},
    author = {Nguyen, P D H and Hoffmann, M and Pattacini, U and Metta, G},
    month = {8},
    pages = {163--170},
    doi = {10.1109/DEVLRN.2019.8850681}
}

@inproceedings{pathak19disagreement,
    title = {{Self-Supervised Exploration via Disagreement}},
    year = {2019},
    booktitle = {International Conference on Machine Learning (ICML)},
    author = {Pathak, Deepak and Gandhi, Dhiraj and Gupta, Abhinav}
}

@inproceedings{Schaul2015_UVFA,
    title = {{Universal Value Function Approximators}},
    year = {2015},
    booktitle = {International Conference on Machine Learning (ICML)},
    author = {Schaul, Tom and Horgan, Dan and Gregor, Karol and Silver, David},
    pages = {1312–1320},
    url = {http://proceedings.mlr.press/v37/schaul15.pdf}
}

@inproceedings{Lang2018ARobots,
    title = {{A deep convolutional neural network model for sense of agency and object permanence in robots}},
    year = {2018},
    booktitle = {ICDL-EpiRob 2018},
    author = {Lang, Claus and Schillaci, Guido and Hafner, Verena V},
    pages = {257--262},
    publisher = {IEEE},
    isbn = {9781538661109},
    doi = {10.1109/DEVLRN.2018.8761015}
}

@inproceedings{Mnih2016AsynchronousLearningb,
    title = {{Asynchronous methods for deep reinforcement learning}},
    year = {2016},
    booktitle = {International Conference on Machine Learning (ICML)},
    author = {Mnih, Volodymyr and Badia, Adria Puigdomenech and Mirza, Lehdi and Graves, Alex and Harley, Tim and Lillicrap, Timothy P and Silver, David and Kavukcuoglu, Koray},
    pages = {2850--2869},
    volume = {4},
    publisher = {PMLR},
    url = {http://proceedings.mlr.press/v48/mniha16.html},
    isbn = {9781510829008},
    issn = {1938-7228},
    arxivId = {1602.01783}
}

@inproceedings{Roncone2014AutomaticRobot,
    title = {{Automatic kinematic chain calibration using artificial skin: Self-touch in the iCub humanoid robot}},
    year = {2014},
    booktitle = {IEEE International Conference on Robotics and Automation},
    author = {Roncone, Alessandro and Hoffmann, Matej and Pattacini, Ugo and Metta, Giorgio},
    pages = {2305--2312},
    isbn = {9781479936847},
    doi = {10.1109/ICRA.2014.6907178},
    issn = {10504729}
}

@article{deVignemont2010BodyCons,
    title = {{Body schema and body image-Pros and cons}},
    year = {2010},
    journal = {Neuropsychologia},
    author = {de Vignemont, Frederique},
    doi = {10.1016/j.neuropsychologia.2009.09.022},
    issn = {00283932}
}

@article{Pugach2019Brain-inspiredEvents,
    title = {{Brain-inspired coding of robot body schema through visuo-motor integration of touched events}},
    year = {2019},
    journal = {Frontiers in Neurorobotics},
    author = {Pugach, Ganna and Pitti, Alexandre and Tolochko, Olga and Gaussier, Philippe},
    number = {March},
    volume = {13},
    doi = {10.3389/fnbot.2019.00005},
    issn = {16625218}
}

@inproceedings{Nguyen2018CompactInteraction,
    title = {{Compact Real-time Avoidance on a Humanoid Robot for Human-robot Interaction}},
    year = {2018},
    booktitle = {ACM/IEEE International Conference on Human-Robot Interaction},
    author = {Nguyen, D.H.P. and Hoffmann, M. and Roncone, A. and Pattacini, U. and Metta, G.},
    number = {},
    volume = {},
    isbn = {9781450349536},
    doi = {10.1145/3171221.3171245},
    issn = {21672148}
}

@inproceedings{Lillicrap2015ContinuousLearning,
    title = {{Continuous control with deep reinforcement learning}},
    year = {2015},
    booktitle = {International Conference on Learning Representations (ICLR)},
    author = {Lillicrap, Timothy P. and Hunt, Jonathan J. and Pritzel, Alexander and Heess, Nicolas and Erez, Tom and Tassa, Yuval and Silver, David and Wierstra, Daan},
    url = {http://arxiv.org/abs/1509.02971},
    arxivId = {1509.02971}
}

@article{Hwang2020DealingFramework,
    title = {{Dealing with Large-Scale Spatio-Temporal Patterns in Imitative Interaction between a Robot and a Human by Using the Predictive Coding Framework}},
    year = {2020},
    journal = {IEEE Transactions on Systems, Man, and Cybernetics: Systems},
    author = {Hwang, Jungsik and Kim, Jinhyung and Ahmadi, Ahmadreza and Choi, Minkyu and Tani, Jun},
    number = {5},
    pages = {1918--1931},
    volume = {50},
    publisher = {IEEE},
    doi = {10.1109/TSMC.2018.2791984},
    issn = {21682232}
}

@inproceedings{Silver2014DeterministicAlgorithms,
    title = {{Deterministic Policy Gradient Algorithms}},
    year = {2014},
    booktitle = {International Conference on Machine Learning (ICML)},
    author = {Silver, David and Lever, Guy and Hees, Nicolas and Degris, Thomas and Wierstra, Daan and Riedmiller, Martin},
    pages = {1--9},
    url = {http://proceedings.mlr.press/v32/silver14.pdf%0Apapers://d471b97a-e92c-44c2-8562-4efc271c8c1b/Paper/p652},
    isbn = {9781634393973},
    issn = {1938-7228}
}

@inproceedings{Hinz2018DriftingRobot,
    title = {{Drifting perceptual patterns suggest prediction errors fusion rather than hypothesis selection: Replicating the rubber-hand illusion on a robot}},
    year = {2018},
    booktitle = {ICDL-EpiRob 2018},
    author = {Hinz, Nina Alisa and Lanillos, Pablo and Mueller, Hermann and Cheng, Gordon},
    pages = {125--132},
    publisher = {IEEE},
    isbn = {9781538661109},
    doi = {10.1109/DEVLRN.2018.8761005},
    arxivId = {1806.06809}
}

@article{Elsner2001EffectControl,
    title = {{Effect anticipation and action control}},
    year = {2001},
    journal = {Journal of Experimental Psychology: Human Perception and Performance},
    author = {Elsner, Birgit and Hommel, Bernhard},
    number = {1},
    pages = {229--240},
    volume = {27},
    doi = {10.1037/0096-1523.27.1.229},
    issn = {00961523}
}

@inproceedings{Watter2015EmbedImagesb,
    title = {{Embed to control: A locally linear latent dynamics model for control from raw images}},
    year = {2015},
    booktitle = {Advances in neural information processing systems},
    author = {Watter, Manuel and Springenberg, Jost Tobias and Boedecker, Joschka and Riedmiller, Martin A.},
    pages = {2746--2754},
    doi = {10.1353/cj.0.0057},
    issn = {00097101}
}

@article{Dilokthanakul2019FeatureLearning,
    title = {{Feature Control as Intrinsic Motivation for Hierarchical Reinforcement Learning}},
    year = {2019},
    journal = {IEEE Transactions on Neural Networks and Learning Systems},
    author = {Dilokthanakul, Nat and Kaplanis, Christos and Pawlowski, Nick and Shanahan, Murray},
    doi = {10.1109/TNNLS.2019.2891792},
    issn = {21622388},
    pmid = {30714933},
    arxivId = {1705.06769}
}

@article{Chinellato2011ImplicitReaching,
    title = {{Implicit sensorimotor mapping of the peripersonal space by gazing and reaching}},
    year = {2011},
    journal = {IEEE Transactions on Autonomous Mental Development},
    author = {Chinellato, Eris and Antonelli, Marco and Grzyb, Beata J. and Del Pobil, Angel P.},
    number = {1},
    pages = {43--53},
    volume = {3},
    doi = {10.1109/TAMD.2011.2106781},
    issn = {19430604}
}

@article{Park2018LearningImitate,
    title = {{Learning for Goal-Directed Actions Using RNNPB: Developmental Change of 'What to Imitate'}},
    year = {2018},
    journal = {IEEE Transactions on Cognitive and Developmental Systems},
    author = {Park, Jun Cheol and Kim, Dae Shik and Nagai, Yukie},
    number = {3},
    pages = {545--556},
    volume = {10},
    publisher = {IEEE},
    doi = {10.1109/TCDS.2017.2679765},
    issn = {23798939}
}

@inproceedings{Qureshi2019MotionNetworks,
    title = {{Motion planning networks}},
    year = {2019},
    booktitle = {IEEE International Conference on Robotics and Automation},
    author = {Qureshi, Ahmed H. and Simeonov, Anthony and Bency, Mayur J. and Yip, Michael C.},
    pages = {2118--2124},
    publisher = {IEEE},
    isbn = {9781538660263},
    doi = {10.1109/ICRA.2019.8793889},
    issn = {10504729},
    arxivId = {1806.05767}
}

@inproceedings{Copete2017MotorLearning,
    title = {{Motor development facilitates the prediction of others' actions through sensorimotor predictive learning}},
    year = {2017},
    booktitle = {ICDL-EpiRob 2016},
    author = {Copete, Jorge Luis and Nagai, Yukie and Asada, Minoru},
    pages = {223--229},
    publisher = {IEEE},
    isbn = {9781509050697},
    doi = {10.1109/DEVLRN.2016.7846823}
}

@inproceedings{Todorov2012MuJoCo:Control,
    title = {{MuJoCo: A physics engine for model-based control}},
    year = {2012},
    booktitle = {IEEE International Conference on Intelligent Robots and Systems},
    author = {Todorov, Emanuel and Erez, Tom and Tassa, Yuval},
    pages = {5026--5033},
    isbn = {9781467317375},
    doi = {10.1109/IROS.2012.6386109},
    issn = {21530858}
}

@book{Bremner2012MultisensoryDevelopment,
    title = {{Multisensory Development}},
    year = {2012},
    author = {Bremner, Andrew J. and Lewknowicz, David J. and Spence, Charles},
    pages = {392},
    publisher = {Oxford University Press},
    url = {https://www.oxfordscholarship.com/10.1093/acprof:oso/9780199586059.001.0001/acprof-9780199586059},
    address = {Oxford},
    isbn = {9780199586059},
    doi = {10.1093/acprof:oso/9780199586059.001.0001},
    language = {eng}
}

@article{Vicente2016OnlineFeedback,
    title = {{Online body schema adaptation based on internal mental simulation and multisensory feedback}},
    year = {2016},
    journal = {Frontiers Robotics AI},
    author = {Vicente, Pedro and Jamone, Lorenzo and Bernardino, Alexandre},
    number = {MAR},
    volume = {3},
    doi = {10.3389/frobt.2016.00007},
    issn = {22969144}
}

@inproceedings{Schillaci2014OnlineModel,
    title = {{Online learning of visuo-motor coordination in a humanoid robot. A biologically inspired model}},
    year = {2014},
    booktitle = {ICDL-EPIROB 2014},
    author = {Schillaci, Guido and Hafner, Verena V. and Lara, Bruno},
    pages = {130--136},
    publisher = {IEEE},
    isbn = {9781479975402},
    doi = {10.1109/DEVLRN.2014.6982967}
}

@article{Serino2019PeripersonalSelf,
    title = {{Peripersonal space (PPS) as a multisensory interface between the individual and the environment, defining the space of the self}},
    year = {2019},
    journal = {Neuroscience and Biobehavioral Reviews},
    author = {Serino, Andrea},
    number = {August 2018},
    pages = {138--159},
    volume = {99},
    publisher = {Elsevier},
    url = {https://doi.org/10.1016/j.neubiorev.2019.01.016},
    doi = {10.1016/j.neubiorev.2019.01.016},
    issn = {18737528}
}

@article{Roncone2016PeripersonalSkin,
    title = {{Peripersonal space and margin of safety around the body: Learning visuo-tactile associations in a humanoid robot with artificial skin}},
    year = {2016},
    journal = {PLoS ONE},
    author = {Roncone, Alessandro and Hoffmann, Matej and Pattacini, Ugo and Fadiga, Luciano and Metta, Giorgio},
    number = {10},
    pages = {1--32},
    volume = {11},
    doi = {10.1371/journal.pone.0163713},
    issn = {19326203}
}

@article{Wijesinghe2018RobotIntegration,
    title = {{Robot end effector tracking using predictive multisensory integration}},
    year = {2018},
    journal = {Frontiers in Neurorobotics},
    author = {Wijesinghe, Lakshitha P. and Triesch, Jochen and Shi, Bertram E.},
    number = {October},
    pages = {1--16},
    volume = {12},
    doi = {10.3389/fnbot.2018.00066},
    issn = {16625218}
}

@inproceedings{Lanillos2020RobotMirror,
    title = {{Robot self/other distinction: active inference meets neural networks learning in a mirror}},
    year = {2020},
    booktitle = {24th European Conference on Artificial Intelligence (ECAI 2020)},
    author = {Lanillos, Pablo and Pages, Jordi and Cheng, Gordon},
    number = {January},
    url = {http://arxiv.org/abs/2004.05473},
    isbn = {9781643681009},
    doi = {10.3233/FAIA200372},
    issn = {09226389},
    arxivId = {2004.05473}
}

@article{Hoffmann2018RoboticCortex,
    title = {{Robotic homunculus: Learning of artificial skin representation in a humanoid robot motivated by primary somatosensory cortex}},
    year = {2018},
    journal = {IEEE Transactions on Cognitive and Developmental Systems},
    author = {Hoffmann, Matej and Straka, Zdenek and Farkas, Igor and Vavrecka, Michal and Metta, Giorgio},
    number = {2},
    pages = {163--176},
    volume = {10},
    doi = {10.1109/TCDS.2017.2649225},
    issn = {23798939}
}

@inproceedings{Byravan2018SE3-Pose-Nets:Control,
    title = {{SE3-Pose-Nets: Structured deep dynamics models for visuomotor control}},
    year = {2018},
    booktitle = {IEEE International Conference on Robotics and Automation},
    author = {Byravan, Arunkumar and Lceb, Felix and Meier, Franziska and Fox, Dieter},
    pages = {3339--3346},
    isbn = {9781538630815},
    doi = {10.1109/ICRA.2018.8461184},
    issn = {10504729},
    arxivId = {1710.00489}
}

@article{Verschoor2017Self-by-doing:Self-acquisition,
    title = {{Self-by-doing: The role of action for self-acquisition}},
    year = {2017},
    journal = {Social Cognition},
    author = {Verschoor, Stephan A. and Hommel, Bernhard},
    number = {2},
    pages = {127--145},
    volume = {35},
    doi = {10.1521/soco.2017.35.2.127},
    issn = {0278016X}
}

@article{Rochat1995SpatialInfants,
    title = {{Spatial Determinants in the Perception of Self-Produced Leg Movements by 3- to 5-Month-Old Infants}},
    year = {1995},
    journal = {Developmental Psychology},
    author = {Rochat, Philippe and Morgan, Rachel},
    number = {4},
    pages = {626--636},
    volume = {31},
    doi = {10.1037/0012-1649.31.4.626},
    issn = {00121649}
}

@article{Gallese2010TheAction,
    title = {{The bodily self as power for action}},
    year = {2010},
    journal = {Neuropsychologia},
    author = {Gallese, Vittorio and Sinigaglia, Corrado},
    number = {3},
    pages = {746--755},
    volume = {48},
    doi = {10.1016/j.neuropsychologia.2009.09.038},
    issn = {00283932},
    pmid = {19835895}
}

@article{Corbetta2018TheCoordination,
    title = {{The Embodied Origins of Infant Reaching: Implications for the Emergence of Eye-Hand Coordination}},
    year = {2018},
    journal = {Kinesiology Review},
    author = {Corbetta, Daniela and Wiener, Rebecca F. and Thurman, Sabrina L. and McMahon, Emalie},
    number = {1},
    pages = {10--17},
    volume = {7},
    doi = {10.1123/kr.2017-0052},
    issn = {2163-0453}
}

@article{Maravita2004Toolsschema,
    title = {{Tools for the body (schema)}},
    year = {2004},
    journal = {Trends in Cognitive Sciences},
    author = {Maravita, Angelo and Iriki, Atsushi},
    number = {2},
    pages = {79--86},
    volume = {8},
    doi = {10.1016/j.tics.2003.12.008},
    issn = {13646613},
    pmid = {15588812}
}

@inproceedings{Nguyen2018TransferringTasks,
    title = {{Transferring Visuomotor Learning from Simulation to the Real World for Robotics Manipulation Tasks}},
    year = {2018},
    booktitle = {2018 IEEE/RSJ International Conference on Intelligent Robots and Systems (IROS)},
    author = {Nguyen, Phuong D.H. and Fischer, Tobias and Chang, Hyung Jin and Pattacini, Ugo and Metta, Giorgio and Demiris, Yiannis},
    month = {10},
    pages = {6667--6674},
    publisher = {IEEE},
    url = {https://ieeexplore.ieee.org/document/8594519/},
    isbn = {978-1-5386-8094-0},
    doi = {10.1109/IROS.2018.8594519}
}

@inproceedings{Srinivas2018UniversalNetworks,
    title = {{Universal planning networks}},
    year = {2018},
    booktitle = {International Conference on Machine Learning (ICML)},
    author = {Srinivas, Aravind and Jabri, Allan and Abbeel, Pieter and Levine, Sergey and Finn, Chelsea},
    volume = {11},
    isbn = {9781510867963},
    arxivId = {1804.00645}
}

@article{Lanillos2017YieldingContingencies,
    title = {{Yielding Self-Perception in Robots Through Sensorimotor Contingencies}},
    year = {2017},
    journal = {IEEE Transactions on Cognitive and Developmental Systems},
    author = {Lanillos, Pablo and Dean-Leon, Emmanuel and Cheng, Gordon},
    number = {2},
    pages = {100--112},
    volume = {9},
    doi = {10.1109/TCDS.2016.2627820},
    issn = {23798939}
}

\end{document}